%% file: main_phys_rev.tex
\documentclass[%
 reprint,
 amsmath,amssymb,
 aps,
]{revtex4-2}

\usepackage{graphicx}
\usepackage{dcolumn}
\usepackage{bm}
\usepackage{hyperref}

\usepackage{caption}
\usepackage{subcaption}
\usepackage[]{algorithm2e}

\begin{document}

\preprint{APS/123-QED}

\newcommand\note[1]{\textcolor{red}{\large \textbf{NOTE: #1}}}
\newcommand{\argmin}
{\mathrm{arg}\operatornamewithlimits{\,min}}
\newcommand{\argmax}{\mathrm{arg}\operatornamewithlimits{\,max}}
\newcommand{\Lcal}{\mathcal{L}}

\newcommand{\G}{G_{\theta}}
\newcommand{\D}{D_{\phi}}
\newcommand{\E}{E_{\tau}}

\newcommand{\pg}{\theta}
\newcommand{\pd}{\phi}
\newcommand{\pe}{\tau}

\title{Reconstruction of 3D Porous Media From 2D Slices}


\author{Denis Volkhonskiy}
\email{Denis.Volkhonskiy@skoltech.ru}
\affiliation{Skolkovo Institute of Science and Technology}

\author{Ekaterina Muravleva}
\email{E.Muravleva@skoltech.ru}
\altaffiliation[Also at ]{Digital Petroleum LLC}
\affiliation{Skolkovo Institute of Science and Technology}

\author{Oleg Sudakov}
\affiliation{Skolkovo Institute of Science and Technology}

\author{Denis Orlov}
\altaffiliation[Also at ]{Digital Petroleum LLC}
\affiliation{Skolkovo Institute of Science and Technology}

\author{Boris Belozerov}
\author{Vladislav Krutko}
\affiliation{%
 Gazprom Neft Science \& Technology Center
}%
\author{Evgeny Burnaev}
\affiliation{Skolkovo Institute of Science and Technology}

\author{Dmitry Koroteev}
\altaffiliation[Also at ]{Digital Petroleum LLC}
\affiliation{Skolkovo Institute of Science and Technology}


\begin{abstract}
In many branches of earth sciences, the problem of rock study on the micro-level arises.  However, a significant number of representative samples is not always feasible. Thus the problem of the generation of samples with similar properties becomes actual. In this paper, we propose a novel deep learning architecture for three-dimensional porous media reconstruction from two-dimensional slices. We fit a distribution on all possible three-dimensional structures of a specific type based on the given dataset of samples. Then, given partial information (central slices), we recover the three-dimensional structure around such slices as the most probable one according to that constructed distribution. Technically, we implement this in the form of a deep neural network with encoder, generator and discriminator modules. Numerical experiments show that this method provides a good reconstruction in terms of Minkowski functionals.
\end{abstract}

\maketitle


\input{sections_phys_rev/1_intro}

\input{sections_phys_rev/2_related}
\input{sections_phys_rev/3_method}

\input{sections_phys_rev/4_experiments}
\input{sections_phys_rev/5_conclusion}

\section*{Acknowledgements}

The work of D. Koroteev on the the applied problem side and the related tasks was supported by the Ministry of Science and Higher Education of the Russian Federation under agreement No. 075-10-2020-119 within the framework of the development program for a world-class Research Center. The work of E. Burnaev on multiscale neurodynamic generative models was supported by Ministry of Science and Higher Education grant No. 075-10-2021-068.


\clearpage
\bibliography{refs}

\end{document}

%% file: sections_phys_rev/1_intro.tex
\section{Introduction}

Transport processes in soils or other porous media strongly depend on their structure. 
The necessity of modeling such type of processes appears in many practical engineering applications, such as hydrogeology, underground mining, petroleum exploitation,  and contaminant cleanup. 
Digital Rock Physics (DRP) technology \cite{andra2013digital_1, andra2013digital_2, blunt2013pore, koroteev2014direct, orlov2021different, ebadi2021nonlinear} is becoming an essential part of reservoir rock characterization workflow nowadays. Water and hydrocarbon production industries actively use DRP technology for getting insights on the mechanisms of complex fluid movement in a porous space of a reservoir rock \cite{berg2017industrial}, including the rocks with unresolved porous space \cite{ebadi2021strengthening}. The technology aims at the calculation of various physical properties of a rock sample based on its digital representation. The list of properties can include
storage properties such as porosity, open porosity, connected porosity, fractional porosity etc;
transport properties such as permeability, relative phase permeability, capillary pressure;
electromagnetic properties such as formation factor, dielectric permittivity, magnetic and electric permeability etc; elastic coefficients; geomechanical constants; characteristics of NMR response and responses to other good logging signals \cite{evseev2015coupling}.

Digitalization of a rock sample \cite{andra2013digital_1, blunt2013pore, koroteev2017method, chauhan2016processing, sidorenko2021deep} typically covers
 selection of a set of representative core plugs (30mm or 1-inch scale) and drilling out some miniplugs from them (1 to 10 mm scale); X-ray microcomputed tomography ($\mu$CT) scanning of the miniplugs;
processing the grayscale 3D $\mu$CT images for distinguishing between minerals and void pores (segmentation).
In more complicated cases, when the rock samples have the significant amount of pores with submicron sizes, digitalization may be supplemented by 2D imaging of a nm resolution (e.g., with a scanning electron microscope, SEM) for understanding the features of the submicron features. 

The essential part of the Digital Rock Physics efforts is directed towards the characterization of a rather mature oil reservoir (e.g., with an objective to screen the chemical fluids for enhancing oil recovery \cite{koroteev2013application}). In many cases, the core material for these ``old'' reservoirs is lost, but the stored amount of 2D microscopic photos of thin sections is significant. The ability to generate realistic 3D microstructures out of the 2D data enables fulfilling the database of the digitized rocks in the cases when physical core samples are inaccessible. 

In some complicated cases related to submicron porosity, only 2D SEM images might be used to resolve thin channels and pores. The conventional 3D $\mu$CT technique is ineffective here. The tool for reconstruction of 3D data out of 2D sections is of an obvious benefit here as well. We can further use the reconstructed 3D submicron volumes for estimating the physical properties on these pieces of rock enriched with the submicron information. These physical properties can be also estimated using machine learning techniques \cite{sudakov1adriving,muravleva2018application}.

One of the most promising techniques in unsupervised machine learning are generative adversarial networks (GAN) \cite{goodfellowetal2014}, which learn complex probability distributions directly from samples. The first paper on using GANs in the context of 3D porous media generation \cite{mosser2017reconstruction} considered the the task  of generating synthetic images. But no additional information (such as 2D slices) is used in the generation step. 

We propose a new GAN-based deep neural network architecture that can efficiently generate 3D structures given some of the slices of the original image.  This is achieved by introduction of autoencoder module into the deep neural network architecture.

%% file: sections_phys_rev/2_related.tex
\section{Related work}

\subsection{Reconstruction of three-dimensional porous media using Generative Adversarial Networks}
\label{rec}

The standard approach for the reconstruction of three-dimensional porous media consists of an application of some probabilistic spatial models \cite{MPS-review}. The main disadvantages of such methods are the computational time (which may be up to tens of hours per sample) and the regular structure of synthetic samples.

Recently, deep learning methods have become more popular for porous media reconstruction \cite{feng2018accelerating, wang2018porous}. In the first work\cite{mosser2017reconstruction} for the reconstruction of three-dimensional porous media using deep neural networks, the authors proposed to use 3D deep convolutional GAN for 3D porous images. Their model is a standard GAN model \cite{goodfellowetal2014} with 3D convolutional layers, trained on different 3D porous images. For the experiments, they used well-known images of Berea sandstone, Ketton limestone, and Beadpack \cite{imp_college_samples}. The work showed good results in the generation of synthetic 3D images in terms of visual quality and statistical characteristics (permeability, porosity, and Minkowski metrics \cite{minkowski}). In our work, we modify their model in such a way that it can use a 2D slice as an input to generate a 3D image surrounding the given 2D slice.

\subsection{Generative Adversarial Networks}
\label{sec:gans}


Let us describe the main idea of Generative Adversarial Networks (GANs). Consider a dataset of training objects $S_n = \{x_i\}_{i=1}^n$ which are independently sampled from the distribution $p_{data}(x)$. Our goal is to construct a \textit{generator} function $\G(z)$ with parameters $\pg$, where $z$ has a known distribution, and $G(z)$ is distributed as  $p_{data}(x)$. The generator $x = \G(z)$ is a learnable differentiable function of a noise~$z$~from some fixed prior distribution $p_{\mathrm{noise}}(z)$, producing synthetic objects $x$. Let us denote the distribution of the output of the generator (the distribution of the synthetic samples) as $p_{model}(x)$. The goal of the generator is to make $p_{model}(x)$ close to the distribution $p_{data}(x)$ in terms of some metric between probability distributions.

To learn the distribution $p_{model}(x)$ by the generator and make it closer to $p_{data}(x)$, the model requires a second neural network, which is called \textit{discriminator}, $\D(x)$ with parameters $\pd$. This is a standard binary classifier, that distinguishes between two classes: real objects (generated from $p_{data}(x)$) and synthetic objects (generated from $p_{model}(x)$ by the generator $\G(z)$). One can consider the discriminator as a critic that quantifies the quality of the synthetic samples. 

\begin{figure}[t]
  \centering
    \includegraphics[width=8.6cm]{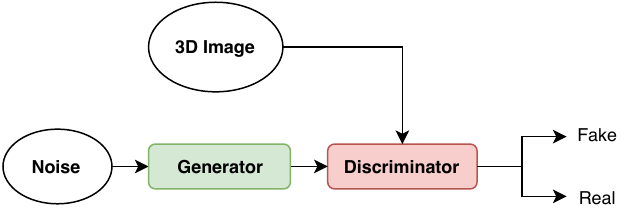}
    \caption{Architecture of the Generative Adversarial Networks}
    \label{fig:gans}
\end{figure}

The complete structure of the generative adversarial framework is presented at Fig. \ref{fig:gans}. 

During training $\G$ tries to generate such realistic images that $\D$ fails in prediction (and confuses whether the object is real or synthetic). On the other hand, $\D$ tries to become accurate to detect the deceiving of $\G$. This leads to the min-max game \eqref{eq:adv_net_function} of a discriminator and a generator 

\begin{multline}
 \label{eq:adv_net_function}
     \underset{\pg}{\min}\,\underset{\pd}{\max}\,\,
        \Lcal(\pg, \pd)  =
          \mathbb{E}_{x\sim p_\mathtt{data}}[\log \D(x)] 
     +\\ \mathbb{E}_{z\sim p_z}[\log(1 - \D(\G(z)))].
\end{multline}
Here $\D(x)$ represents the probability that $x$ comes from the data distribution $p_{data}(x)$ rather than from $p_{model}(x)$. 

The min-max game \eqref{eq:adv_net_function} can be interpreted as follows. First, we should find such discriminator parameters $\pd$, which makes the discriminator as accurate as possible. Then, when we have a reliable discriminator, we train the generator $\G$ (tune its parameters $\pg$) to degrade the discriminator's performance (deceive $\D$). The discriminator $\D$ and the generator $\G$ are realized by deep neural networks with parameters $\pd$ and $\pg$ correspondingly.

In practice, to estimate the value of $\Lcal(\pg, \pd)$ in \eqref{eq:adv_net_function} we approximate both expectations $\mathbb{E}_{x\sim p_\mathtt{data}}$ and $\mathbb{E}_{z\sim p_z}$ by corresponding empirical averages: empirical averaging $\widehat{\mathbb{E}}_{x\sim S_n}$ over the training sample $S_n = \{x_i\}_{i=1}^n$ and empirical averaging $\widehat{\mathbb{E}}_{z\sim p(z)}$ over samples, generated from the prior $p_{noise}(z)$.

We train $\D$ and $\G$ simultaneously. Namely, we make several updates of $\D$ and then several updates of $\G$, etc. Both neural networks are updated via stochastic gradient descent \cite{bottou2010large} or its modifications (in our work, we use ADAM \cite{kingma2014adam} algorithm). The update rules are as follows:

\begin{itemize}
	\item Keeping $\G$ fixed, update $\D$ by ascending its stochastic gradient from equation \eqref{eq:g_update},
    where $\{x_{i_j}\}_{j=1}^k$, $i_j\in\{1,\ldots,n\}$ is a batch of $k$ objects, randomly selected from $S_n$, $\{z_j\}_{j=1}^k$ is a sample from $p_{noise}(z)$, and $k$ is a size of a training batch.
    
	\item Keeping $\D$ fixed, update $\G$ by ascending its stochastic gradient from equation \eqref{eq:d_update}, where $\{z_j\}_{j=1}^k$ is a sample from $p_{noise}(z)$, and $k$ is a size of a training batch.
\end{itemize}

\begin{multline} \label{eq:g_update}
        \frac{\partial}{\partial \pd}\Bigl\{
               \widehat{\mathbb{E}}_{x\sim S_n}[\log \D(x)] +\\ 
               \widehat{\mathbb{E}}_{z\sim p_{noise}(z)}[\log(1 - \D(\G(z)))]
             \Bigr\} = \\
     \frac{\partial}{\partial \pd}\Bigl\{
           \frac{1}{k} \sum\limits_{j=1}^{k} \log \D(x_{i_j}) + \\
           \frac{1}{k} \sum\limits_{j=1}^{k} \log(1 - \D(\G(z_j)))
         \Bigr\} \,,
    \end{multline}
    
\begin{multline} \label{eq:d_update}
\frac{\partial}{\partial \pg}\Bigl\{
      		\widehat{\mathbb{E}}_{z\sim p_{noise}(z)}[\log(1 - \D(\G(z)))] \Bigr\} = \\
\frac{\partial }{\partial \pg}\Bigl\{
      	\frac{1}{k} \sum\limits_{j=1}^{k} \log(1 - \D(\G(z_j))) \Bigr\} \,,
    \end{multline}	


The GANs framework is general, and so it can be applied to data of any type. In this article, we consider each~$x_i$ to be a three-dimensional binary array representing a fixed volume of a 3D porous media.


One of the standard modifications of GANs is conditional Generative Adversarial Networks  (cGANs\cite{mirza2014conditional}). The generator in the original GAN takes as input only the prior noise vector; however, in the case of the cGAN, we also use some prior conditional information as input. In the simplest case, the condition may define the class of the object that we want to generate. 

Since the goal of our work is to propose a model for conditional generation of 3D porous media, given the 2D input slice, we should approximate the distribution $p_{data}(x | s)$, where $x$ is a 3D porous media and $s$ is a input 2D slice, which can be achieved by cGAN.


\subsection{Autoencoders}
\label{auto}

We want to condition the generator on the input 2D slice. However, the black and white 2D slice representation is redundant since it has a specific porous structure and contains large regions of regular shapes, which have the same pixel values (either $0$ or $1$). It leads to the idea of the slice compression before passing it through the generator network.

One of the distinctive features of deep learning models is a good representational learning capability in the unsupervised setting. For this goal \textit{autoencoders} are among the most popular architectures \cite{doersch2016tutorial}. This is a class of models that consists of two neural networks: \textit{encoder} and \textit{decoder}. The encoder takes as input object description and returns its latent representation of a smaller dimension (reduction of object description redundancy). The purpose of the decoder is to transform the latent representation back to the initial description without significant loss of information.

More formally, let us consider the set of training objects $S_n = \{x_i\}_{i=1}^n$. Let us denote the \textit{encoder} with parameters $\theta_E$ as $E_{ae}(x, \theta_E)$ and the \textit{decoder} with parameters $\theta_D$ as $D_{ae}(x, \theta_D)$. If we apply $E_{ae}(x, \theta_E)$ to $x_i$, we obtain the latent representation of the object $h_i = E_{ae}(x_i, \theta_E)$. In order to reconstruct the object from the latent description we apply the decoder and get $\hat{x}_i = D_{ae}(h_i, \theta_D)$. Our aim is to get a latent representation of some fixed dimension, such that $\mathrm{dim}(h)\ll\mathrm{dim}(x)$ and we can recover an object description from its latent representation as accurately as possible. Thus we can formulate the following optimization problem 
\begin{equation}
\label{eq:ae_loss}
    \parallel x_i - D_{ae}(E_{ae}(x_i, \theta_E), \theta_D) \parallel^2_2 \to \min_{\theta_E, \theta_D}.
\end{equation}

During training, the goal is to get the reconstruction error \eqref{eq:ae_loss} as small as possible. For this purpose, we optimize the Euclidean distance between the original input and the reconstruction results. For the optimization, one can use standard stochastic gradient descent or its modifications (for example, ADAM \cite{kingma2014adam}).

%% file: sections_phys_rev/3_method.tex
\section{Method}
\label{sec:spgan}

The goal of this work is to construct a GANs-based framework for 3D porous media synthesis. As we discussed earlier, one of the important requirement is the possibility to take a slice as the input for the generator.

\begin{figure}[t]
  
  \centering
    \includegraphics[width=8.6cm]{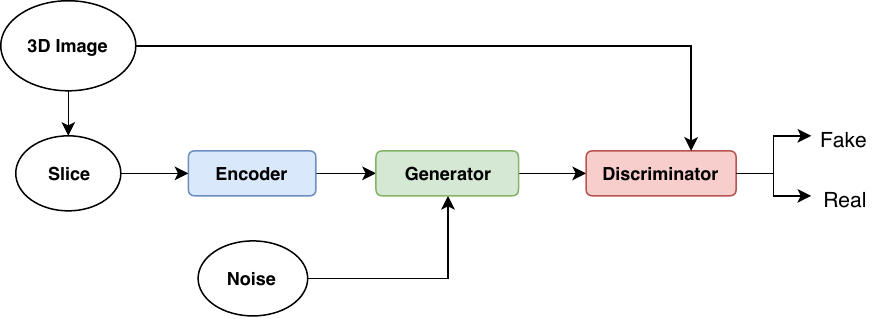}
    \caption{Architecture of Slice to Pores Generative Adversarial Networks}
    \label{fig:sepgan}
\end{figure}

We introduce a new model called Slice to Pores Generative Adversarial Networks (SPGAN). The SPGAN model is a synergy of a convolutional autoencoder and 3D Deep Convolutional Generative Adversarial Networks models. It can be considered as a DCGAN, where the generator is conditioned on the noise and the latent representation of the 2D slice. Our model consists of three neural networks:

\begin{itemize}
\item \textit{Encoder} $\E(s)$ with parameters $\pe$ --- transforms input slice $s$ to a vector representation $h$;

\item \textit{Generator} $\G(z, h)$ with parameters $\pg$ --- transforms the input noise vector $z$ and encoded slice $h$ to a 3D image $x$;

\item \textit{Discriminator} $\D(x)$ with parameters $\pd$ --- predicts the class of input 3D image $x$. This is a standard GANs discriminator.
\end{itemize}
Pair of generator and discriminator is a GAN model. In the next sections we will describe these models separately, and then combine them into the one model, that is presented in Fig. \ref{fig:sepgan}.

\subsection{2D Slice Autoencoder}
\label{sec:sepgan_ae}
In order to learn the autoencoder model for 2D slice images, $l_2$ loss is considered. Generator, in this case, plays the role of the decoder. Generator returns a 3D image as an output, and we would like this image to have the central slice to be close to the input one. There are two main differences from standard 2D convolutional autoencoders: 
\begin{enumerate}
\item Our decoder is a 3D convolutional neural network, thus we should be able to get the central 2D slice from it,
\item Decoder takes as an input not only latent representation but also a noise vector from some prior distribution $p_z(z)$.
\end{enumerate}

To obtain the central slice from the 3D image, we introduce a mask $\textbf{M}$. This is a function, that takes 3D image as an input and returns its central slice. 

\begin{equation}\label{eq:enc_loss}
L(s) = \parallel s - \textbf{M} \odot \G(\E(s), z)\parallel^2_2 \to \min_{\pe, \pg}
\end{equation}

The loss function for our autoencoder is \eqref{eq:enc_loss}. The intuition behind it is the following: we minimize the difference of the input slice $s$ and the central slice $\textbf{M}~\odot~\G(\E(s),~z)$ in the generated 3D image.

\subsection{3D Images Generative Adversarial Network}

For the generation of 3D images, we should be able to define a GAN model. Our GAN model also consists of a generator and discriminator, and there are two important features:

\begin{enumerate}
\item We use 3D convolutional layers, since we work with 3D data;
\item The generator is conditioned on the latent representation of a slice, that is obtained from the autoencoder, described in section \ref{sec:sepgan_ae}.
\end{enumerate}

For training generator and discriminator we use the min-max game \eqref{eq:gans_loss}.

\begin{multline}
\label{eq:gans_loss}
	L(\D, \G)
    	=  \mathbb{E}_{x\sim p_{data}(x)}[\log \D(x)]
    		 +\\ \mathbb{E}_{z\sim p_{noise}(z)}[\log(1 - \D(\G(\E(s), z)))]
    			\\ \to \min_{\pg} \max_{\pd}.
\end{multline}

\subsection{Algorithm}

Since we have two different models (autoencoder and GAN), and the generator takes part in both of them, we should provide an algorithm of simultaneous network training. The formal procedure is presented in Algorithm \ref{al:sepgan}.

\begin{algorithm}[t]
 \For{number of training iterations}{
 Sample minibatch of $k$ 3D images $\{ x_1, \ldots, x_k \}$ from the dataset\;
 Obtain the minibatch of slices $\{s_1 = 
 \textbf{M} \odot x_1, \ldots, s_k = \textbf{M} \odot x_k\}$, using the mask $\textbf{M}$\;
  Sample minibatch of $k$ noise vectors $\{z_1, \ldots, z_k\}$ from the prior distribution $p_z(z)$\;
 
 Update the encoder by ascending its stochastic gradient
 
 $$\nabla_{\pe} \frac{1}{k} \sum_{i=1}^{k} \parallel s_i - \textbf{M} \odot \G(\E(s_i), z_i) \parallel^2_2 \;$$
 
 Update the generator by ascending its stochastic gradient
 
 $$\nabla_{\pg} \frac{1}{k} \sum_{i=1}^{k} \parallel s_i - \textbf{M} \odot \G(\E(s_i), z_i) \parallel^2_2 \;$$
 
 Obtain the minibatch of latent representations $\{h_1 = \E(s_1),~\ldots,~ h_k = \E(s_k)\};$
 
 Update the discriminator by ascending its stochastic gradient
 
 $$
 \nabla_{\pd} \frac{1}{k} \sum_{i=1}^{k} [\log \D(x_i) + \log(1 - \D(\G(z_i, h_i))) ] \;
 $$
 
 Update the generator by descending its stochastic gradient
 
 $$\nabla_{\pg} \frac{1}{k} \sum_{i=1}^{k} \left[\log(1 - \D(\G(z_i, h_i))) \right].$$

 }
 \caption{Algorithm of training SPGAN model}
 \label{al:sepgan}
\end{algorithm}

At each iteration, given a minibatch of 3D images and a minibatch of noise vectors, our algorithm works as the following. At the first step, we update the generator and encoder according to the autoencoder loss \eqref{eq:enc_loss}. At the second step, we update the discriminator and generator, according to the GANs loss \eqref{eq:gans_loss}. We use Adam \cite{kingma2014adam} optimization algorithm as a modification of Stochastic Gradient Descent.

\subsection{Details of the training process}
\label{training}

In our experiments, training of the generative model takes around $10$ hours on a single Tesla V100 GPU. We listed parameters of the training algorithm in Tab. \ref{tab:training_parameters}.

\begin{table}[t]
\centering
\caption{Parameters of the training algorithm}
\label{tab:training_parameters}
\begin{tabular}{|c|c|c|c|}
\hline
                  & Berea    & Ketton   & \begin{tabular}[c]{@{}c@{}}South-Russian\\ sandstone\end{tabular} \\ \hline
Learning rate     & $0.0001$ & $0.0001$ & $0.0001$                                                          \\ \hline
dimension of $z$     & $512$    & $512$    & $512$                                                             \\ \hline
Size of training images & $128^3$   & $128^3$   & $128^3$                                                            \\ \hline
Batch size        & $4$     & $4$     & $4$                                                              \\ \hline
Iterations  & $205000$    & $55500$    & $93100$                                                             \\ \hline
\end{tabular}
\end{table}

%% file: sections_phys_rev/4_experiments.tex
\section{Experiments}
\label{exp}

For the numerical experiments, we chose several training images: Berea sandstone, Ketton limestone (both from Imperial College Collection  \cite{imp_college_samples})  
and in-house sandstone. Our sample belongs to South-Russian geologic formations in Western Siberia. It is a finely medium-grained sandstone from the depth of approximately $1500$ m. We trained the proposed deep learning model on them separately and then compared generated images with the real. 



Initially, we had three 3D images of different structures (sandstones, limestones). These images have the following sizes:
$400^3$ voxels with the size of $3\mu$m for Berea,
$256^3$ voxels with the size of $15.2 \mu$m for Ketton and 
$850^3$ voxels with the size of 3 $\mu$m for South-Russian sandstone,
respectively.

\begin{figure*}[t]
\centering
\includegraphics[width=0.9\textwidth]{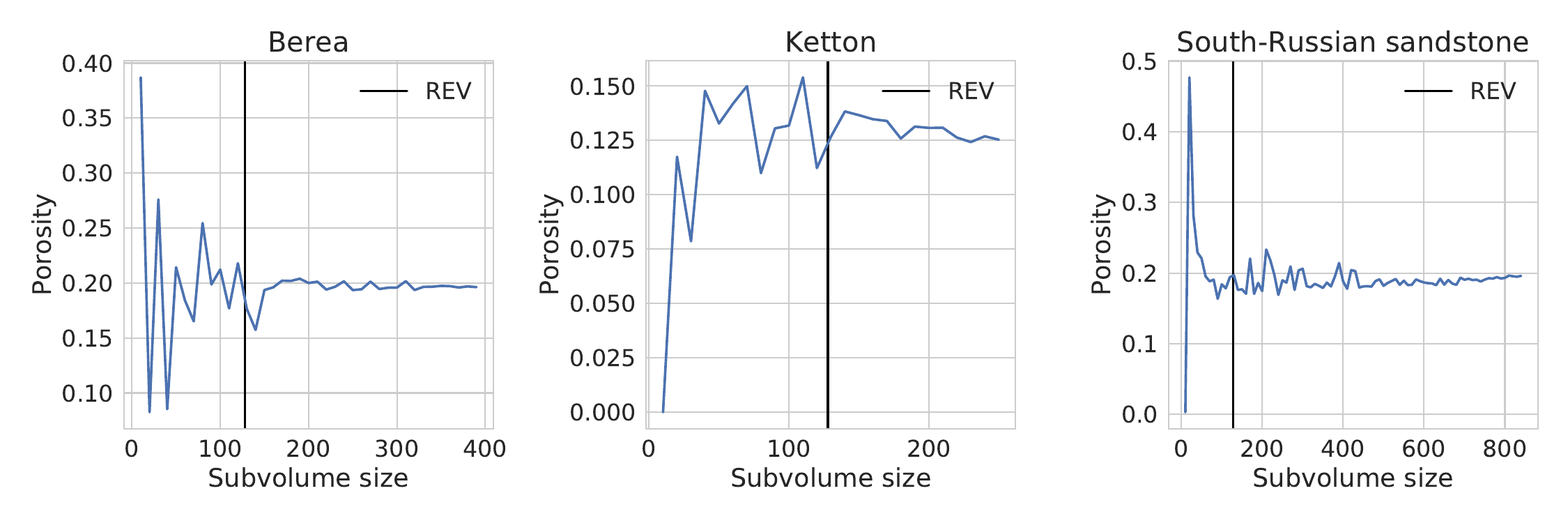}
\caption{Representative Elementary Volume. For each type of porous media we randomly extracted subvolumes. We started from the size of the porous media and decreased it at the size of 10 on each step. We calculated the porosity for each subvolume and determined the Representative Elementary Volume to be $128$.}
\label{fig:rve_porosity}
\end{figure*}

For each type of image, we had only one sample. It is impossible to train a deep neural model for two reasons. First --- we should have a dataset with many samples. Second --- these examples should fit into the GPU memory. To overcome such limitations, on each training iteration we randomly extracted a batch of subvolumes of size $128^3$. 

To select the size of the training samples, we determined a representative elementary volume (REV) based on porosity (see Fig. \ref{fig:rve_porosity}). For all three rock types, REV was more than $120$ voxels. Therefore, for our purposes, subvolume size more than $120$ voxels was an appropriated choice. Thus, we considered cubes of size $128^3$.

\begin{figure}[h!]
    \centering
    \begin{subfigure}[b]{0.2\textwidth}
        \includegraphics[width=\textwidth]{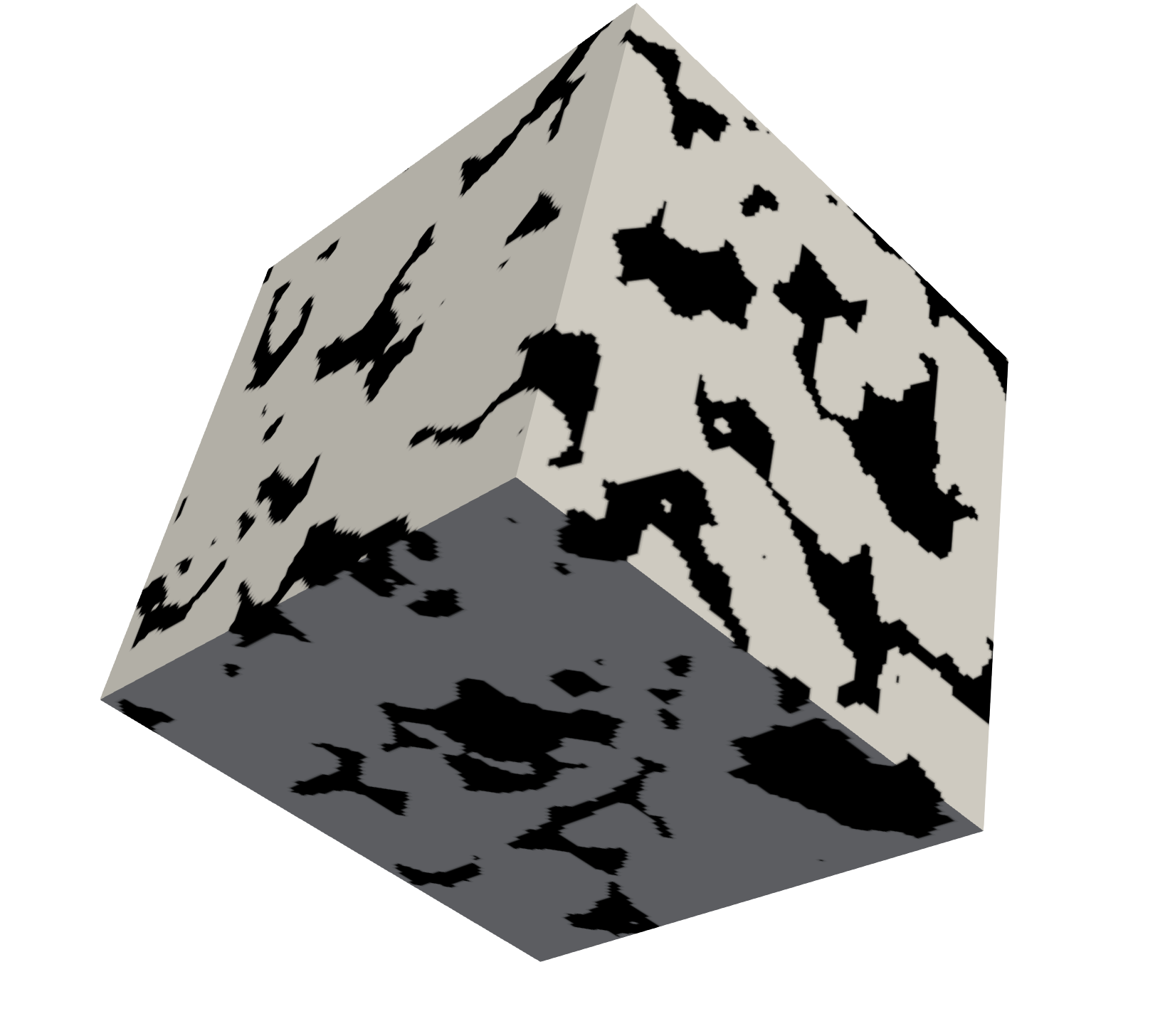}
        \caption{Berea}
    \end{subfigure}
    ~ 
    \begin{subfigure}[b]{0.2\textwidth}
        \includegraphics[width=\textwidth]{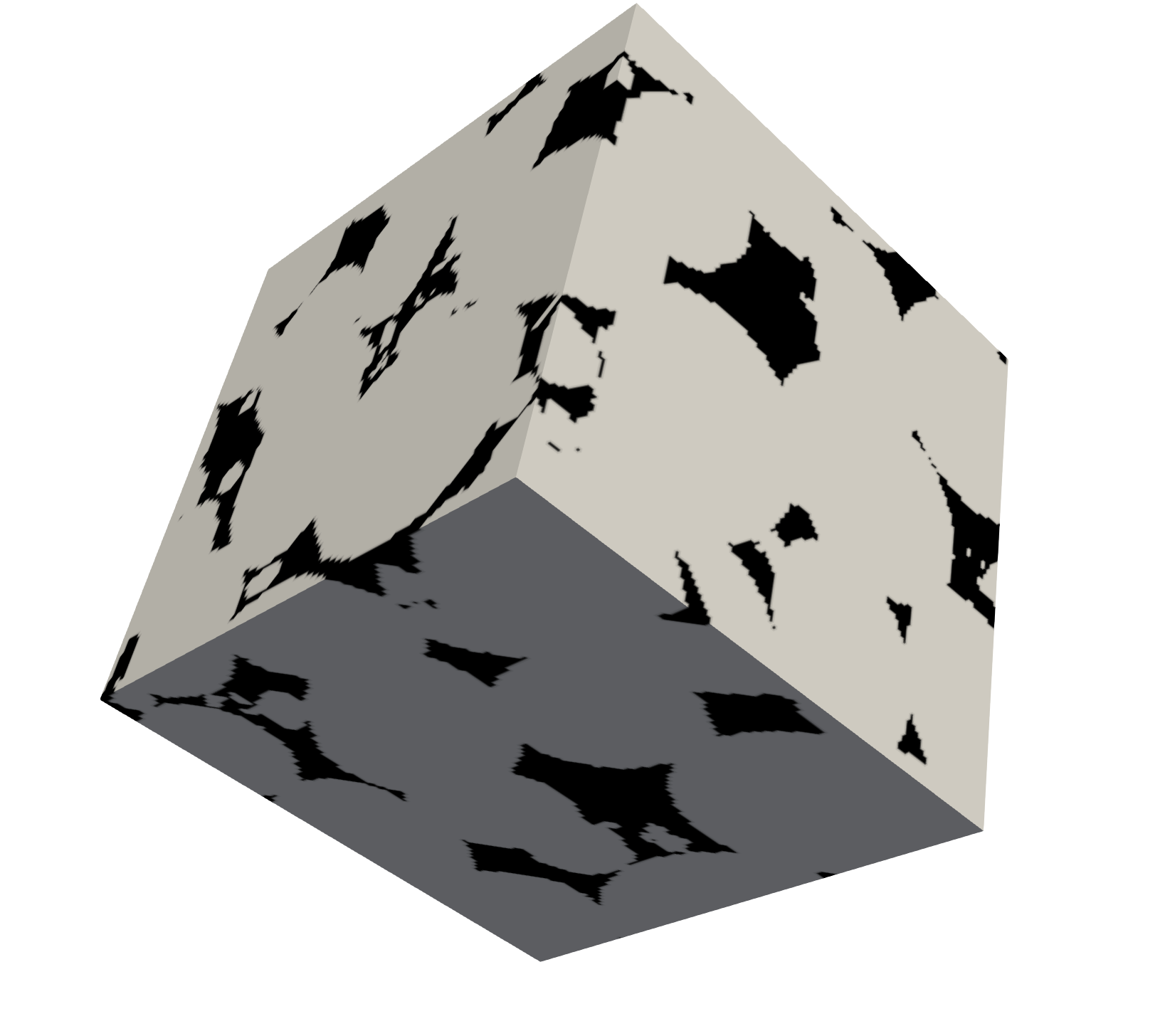}
        \caption{Ketton}
    \end{subfigure}
    ~ 
    \begin{subfigure}[b]{0.2\textwidth}
        \includegraphics[width=\textwidth]{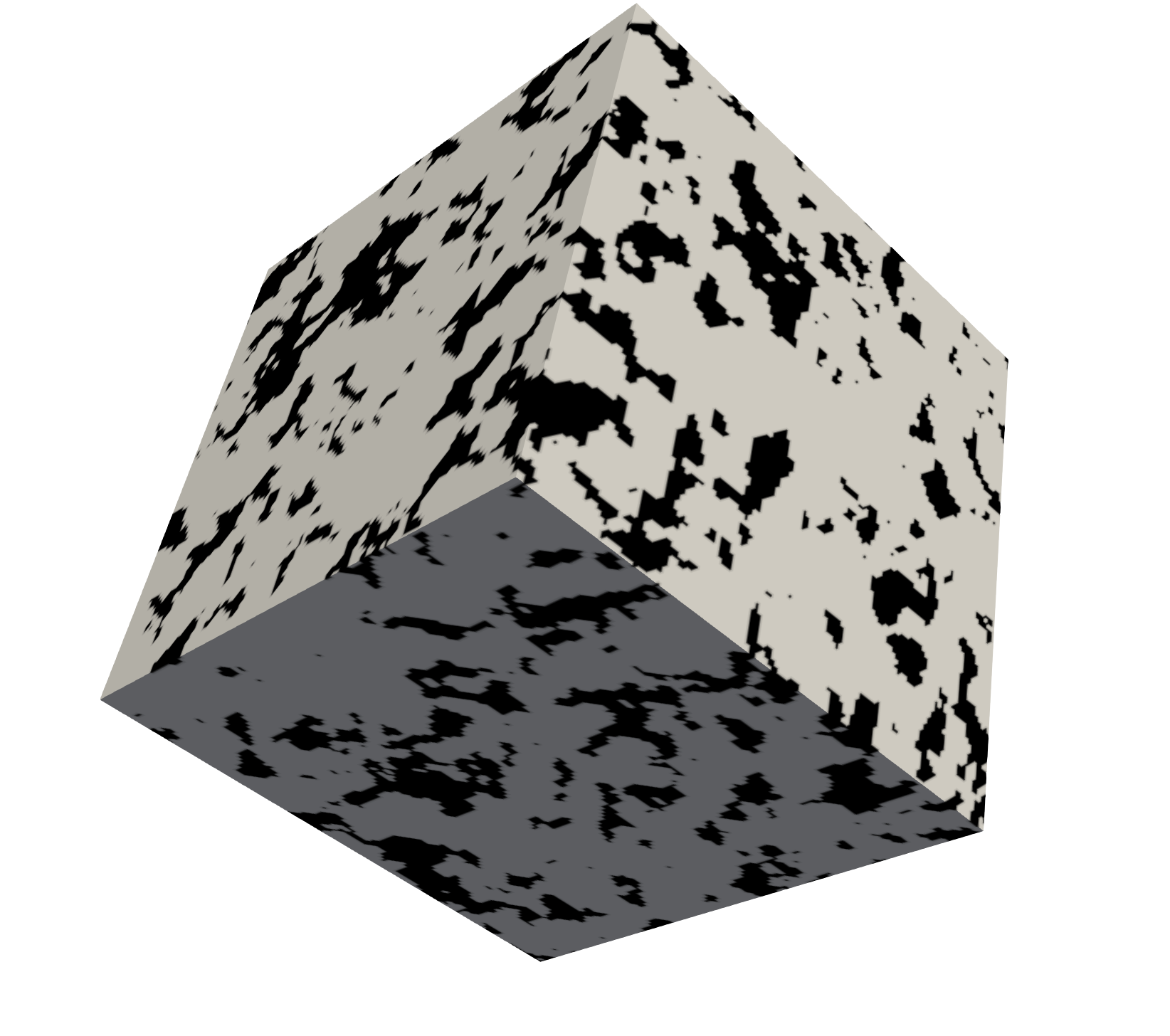}
        \caption{S-R sandstone}
    \end{subfigure}
    \caption{Original 3D samples of three different types: Berea, Ketton, South-Russian sandstone}
    \label{fig:orig3D}
\end{figure}

\begin{figure}[h!]
    \centering
    \begin{subfigure}[b]{0.2\textwidth}
        \includegraphics[width=\textwidth]{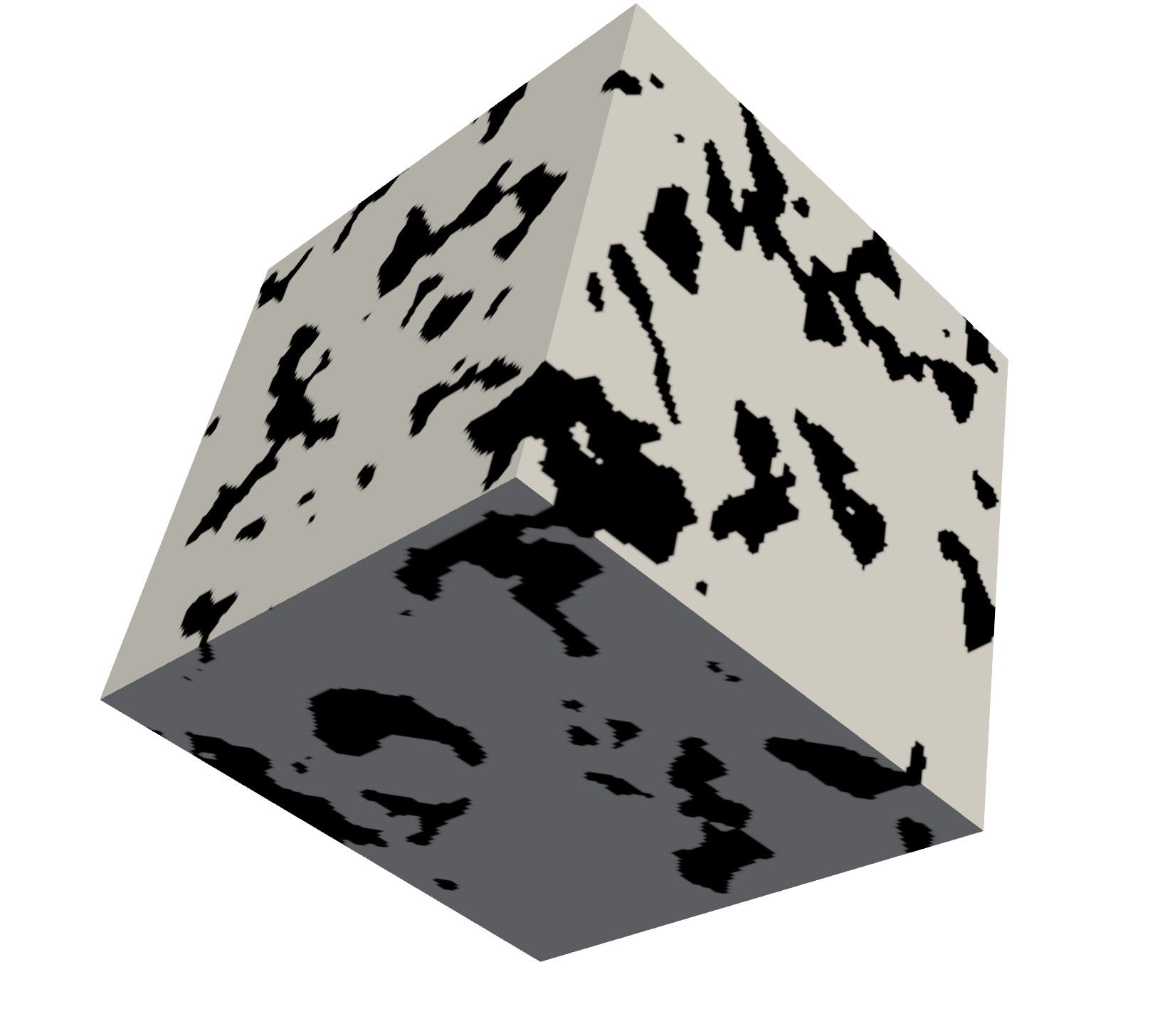}
        \caption{Berea}
    \end{subfigure}
    ~ 
    \begin{subfigure}[b]{0.2\textwidth}
        \includegraphics[width=\textwidth]{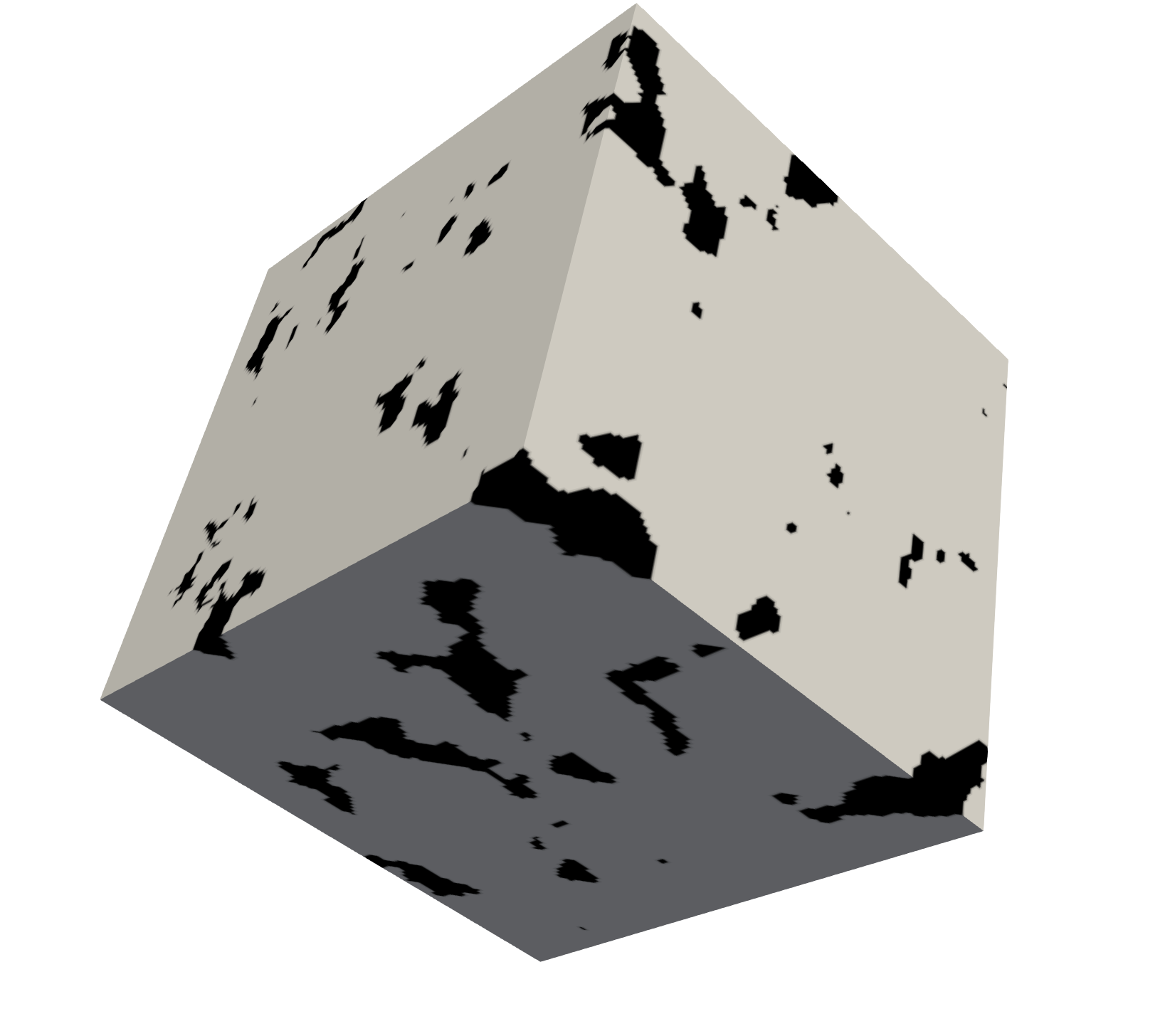}
        \caption{Ketton}
    \end{subfigure}
    ~ 
    \begin{subfigure}[b]{0.2\textwidth}
        \includegraphics[width=\textwidth]{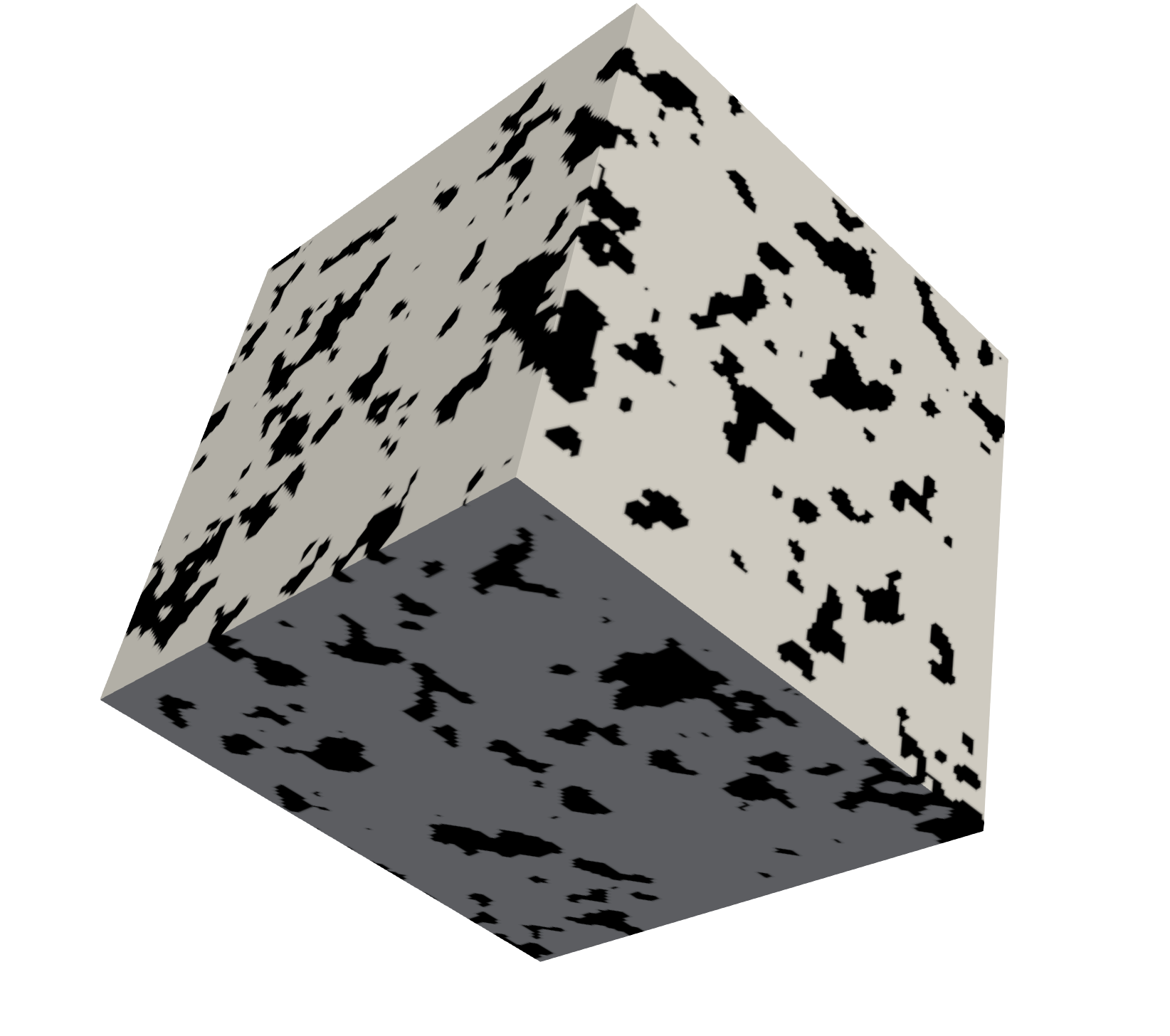}
        \caption{S-R sandstone}
    \end{subfigure}
    \caption{Generated 3D samples of three different types: Berea, Ketton, South-Russian sandstone}
    \label{fig:gen3d}
\end{figure}

3D original and generated samples of size $128^3$ are presented on 
Fig.~\ref{fig:orig3D} and Fig.~\ref{fig:gen3d}, respectively.


One can see that the results obtained by generating different rock samples (Fig. \ref{fig:gen3d}) look visually similar to the original (Fig. \ref{fig:orig3D}). It seems very attractive to quantify complex structures by a limited set of morphological descriptors.  
To measure the quality of synthetic data, we compare porosity and so-called Minkowski\cite{minkowski} functionals. For the comparison, we built box-plots for generated and real samples separately, which is a good visualization of the distribution. The box-plot represents the distribution of the statistics, showing their minimum, maximum, and mean values, along with $25\%$ and $75\%$ quartiles and outliers.

\begin{figure*}[t]
\centering
\includegraphics[width=0.9\textwidth]{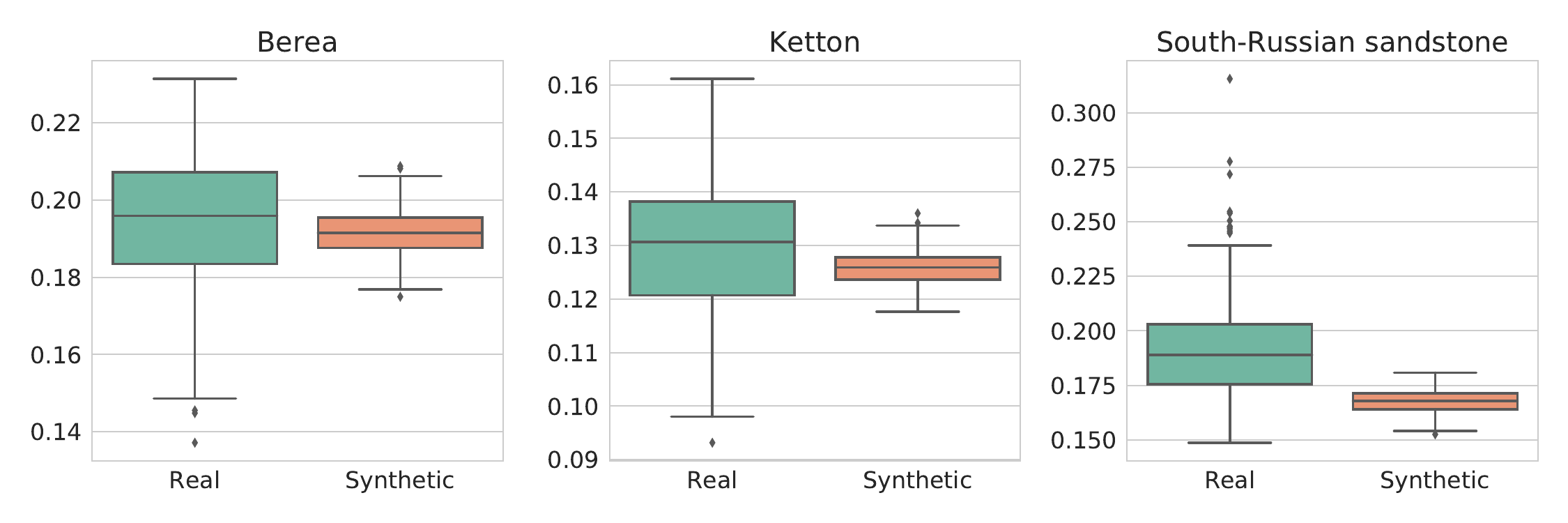}
\caption{Porosity comparison for three types of porous media. Each type is represented by $300$ real and $300$ generated samples. For each sample we computed porosity and created box-plot.}\label{fig:porosity}
\end{figure*}

Porosity is a measure of the void space in the material and is a fraction of the volume of voids over the total volume. It takes values between $0$ and $1$.
The porosity comparison is provided in Fig. \ref{fig:porosity} and shows close agreement.

\begin{figure*}[t]
\centering
\includegraphics[width=0.9\textwidth]{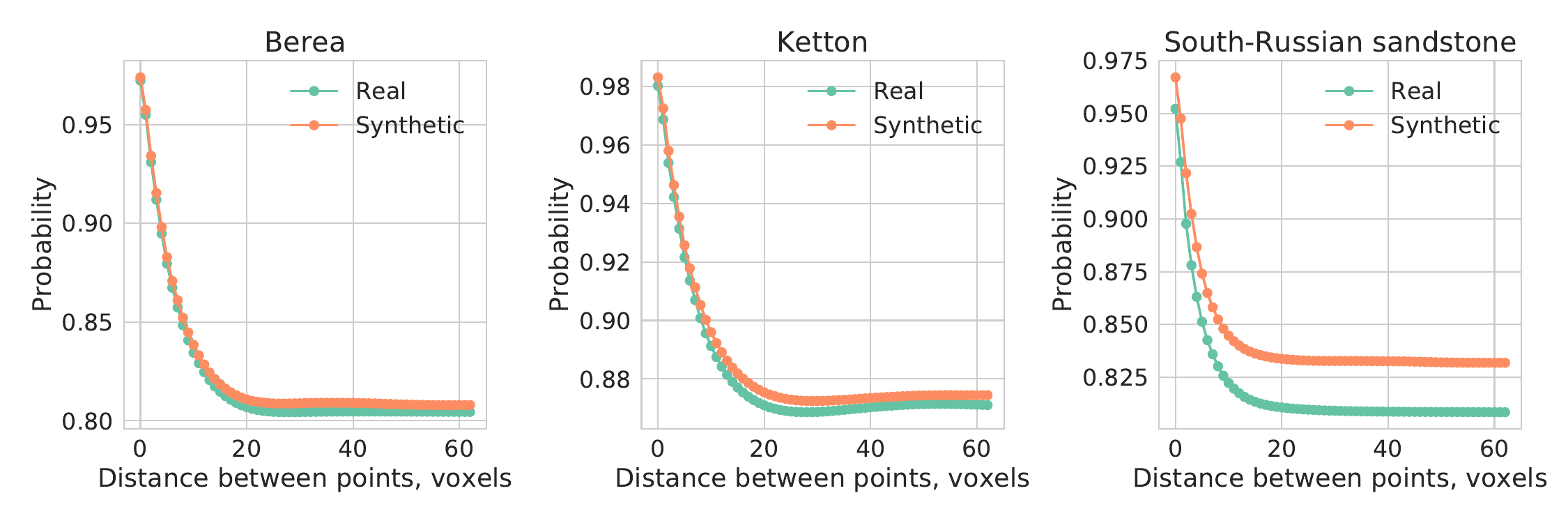}
\caption{Two-Point Correlation Function. For each type of porous media we for both real and synthetic samples we compute probability, that a distance between two points will lie inside the void space. We used PoresPy library \cite{porespy} for computations.}\label{fig:two_point_corr}
\end{figure*}

To analyse pores properties, we computed two-point correlations (TPC  \cite{bazaikin2013numerical}). We used PoresPy library\cite{porespy} code. For two different points, TPC shows the probability of the distance between them lies in the void space. One can mention the small difference between TPC for real and synthetic samples on Fig. \ref{fig:two_point_corr}.

Minkowski functionals are unbiased, stereological estimators and provide local and global morphological information, which are correlated with flow properties \cite{scholz2015direct}. In fact, Minkowski functionals describe the morphology and topology of 2D and 3D binary structures. Let us define the number of vertices as $n_0$, the number of edges as $n_1$, the number of faces as $n_2$, and the number of voxels as $n_3$. Calculation of Minkowski functionals reduces to counting the $n_0,~n_1,~n_2,~n_3$ of the 3D sample. We calculate four Minkowski functionals (of the solid phase): 

\begin{itemize}
    \item Volume 
    
    $$
    V = n_3;
    $$

    \item The surface area (called surface)
    $$
    S = -6 n_3 + 2 . n_2;
    $$
    \item Mean breadth (a quantity proportional to the integral of the mean curvature over the surface)
    
    $$B = 3n_3/2 - n_2 + n_1/2; $$

    \item Euler-Poincare characteristic (connectivity number)
    
    

    $$
    \xi = -n_3 + n_2 - n_1 + n_0.
    $$
\end{itemize}

To compute discrete Minkowski functionals, we used the existing package (MATLAB code) \cite{minkowski}.

\begin{figure*}[h!]
\centering
\includegraphics[width=0.9\textwidth]{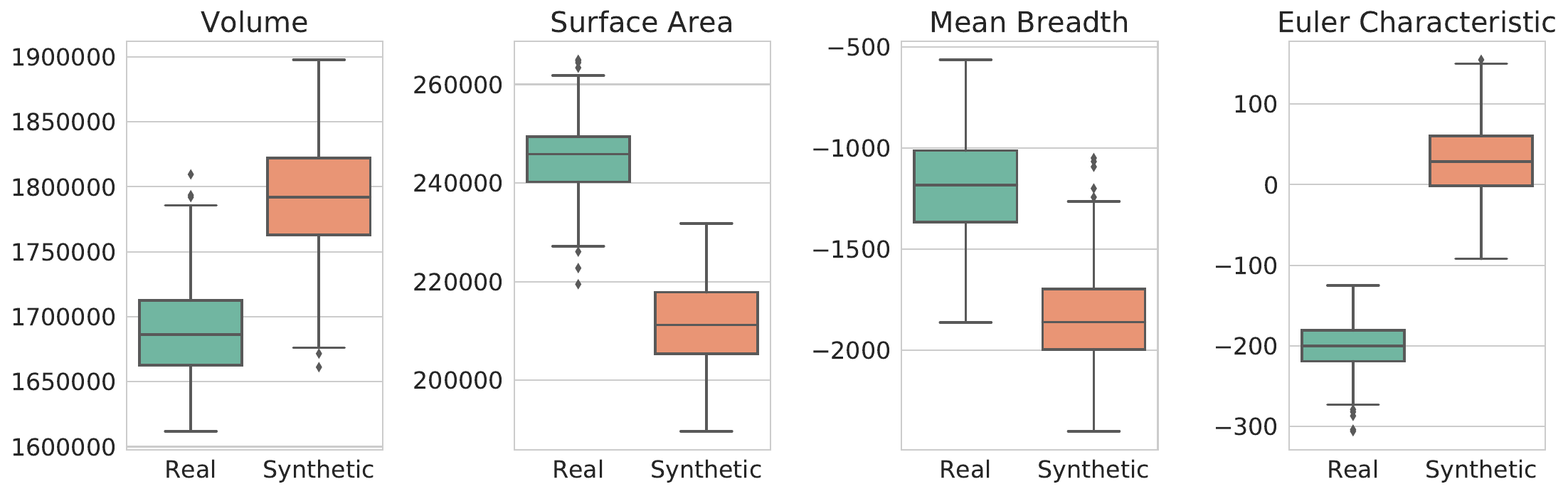}
\caption{Minkowski metrics for Berea: Volume (in voxels), Surface Area (in voxels), Mean Breadth and Euler Characteristics. We took $300$ synthetic samples and $300$ real samples. We computed each Minkowski functional on each sample and compare their distribution using box-plot.}\label{minkowski_berea}
\end{figure*}

\begin{figure*}[h!]
\centering
\includegraphics[width=0.7\textwidth]{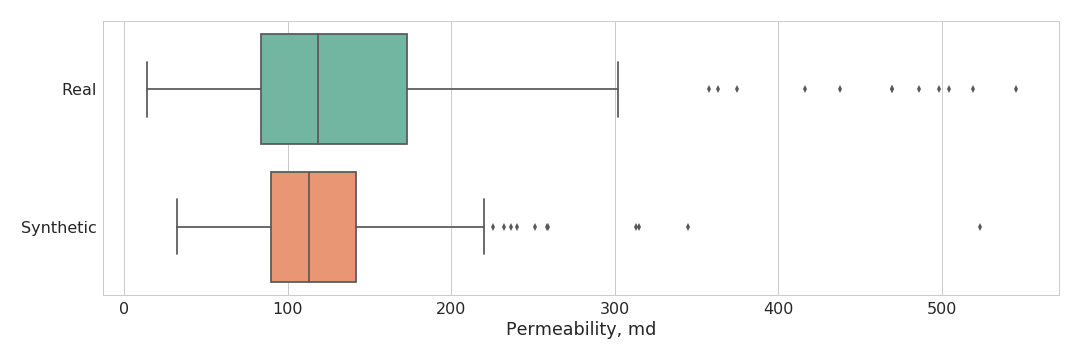}
\caption{Permeability for Berea. We took $300$ synthetic samples and $300$ real samples. For each sample, we computed permeability using Pore Network. We compare two distributions of values using box-plot.}\label{permeability_berea}
\end{figure*}

\begin{figure*}[h!]
\centering
\includegraphics[width=0.9\textwidth]{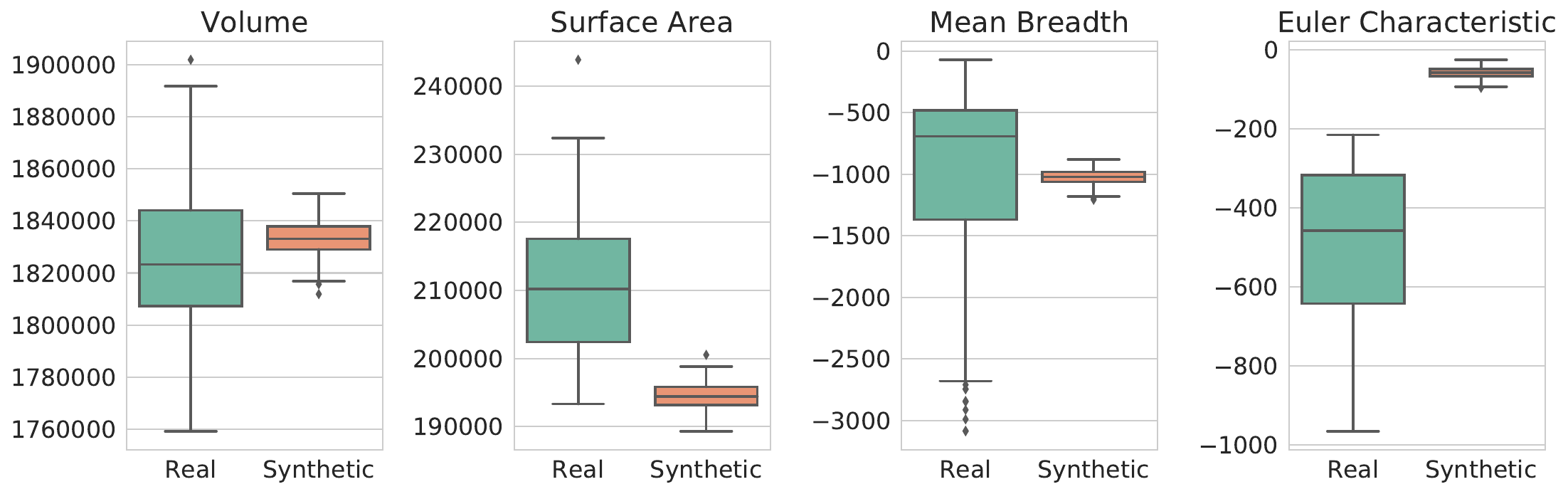}
\caption{Minkowski metrics for Ketton: Volume (in voxels), Surface Area (in voxels), Mean Breadth and Euler Characteristics. We took $300$ synthetic samples and $300$ real samples. We computed each Minkowski functional on each sample and compare their distribution using box-plot.}\label{minkowski_ketton}
\end{figure*}

\begin{figure*}[h!]
\centering
\includegraphics[width=0.7\textwidth]{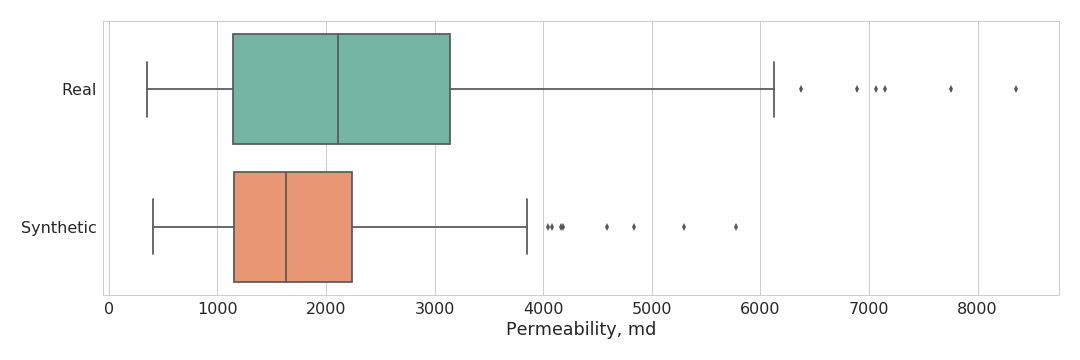}
\caption{Permeability for Ketton. We took $300$ synthetic samples and $300$ real samples. For each sample, we computed permeability using Pore Network. We compare two distributions of values using box-plot.}\label{permeability_ketton}
\end{figure*}

\begin{figure*}[h!]
\centering
\includegraphics[width=0.9\textwidth]{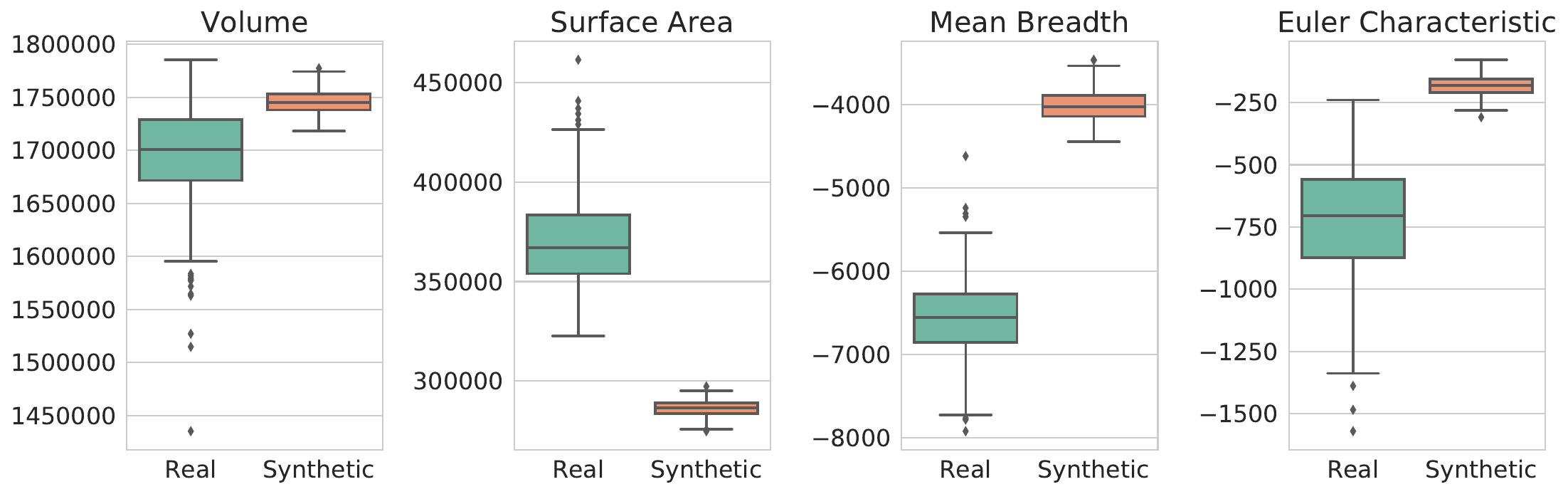}
\caption{Minkowski metrics for South-Russian sandstone: Volume (in voxels), Surface Area (in voxels), Mean Breadth and Euler Characteristics. We took $300$ synthetic samples and $300$ real samples. We computed each Minkowski functional on each sample and compare their distribution using box-plot.}\label{minkowski_ours}
\end{figure*}

\begin{figure*}[h!]
\centering
\includegraphics[width=0.7\textwidth]{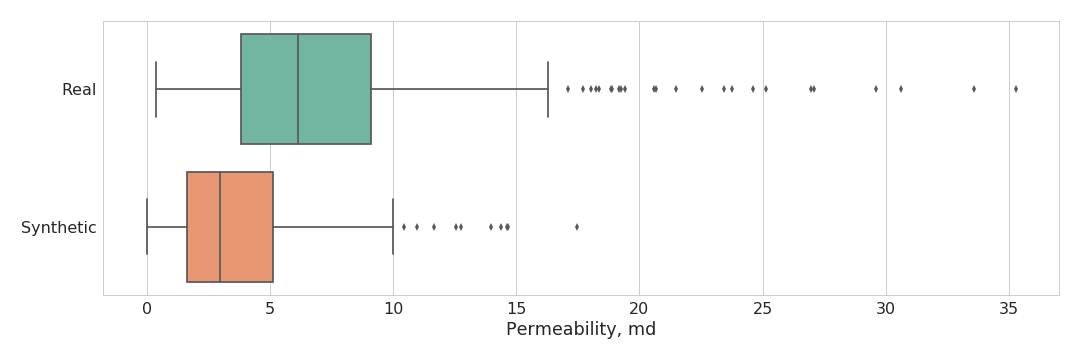}
\caption{Permeability for South-Russian Sandstone. We took $300$ synthetic samples and $300$ real samples. For each sample, we computed permeability using Pore Network. We compare two distributions of values using box-plot.}\label{permeability_ours}
\end{figure*}

The results of comparison with the same metrics for the original data are shown on Fig.~\ref{minkowski_berea} for Berea sandstone, Fig.~\ref{minkowski_ketton} for Ketton limestone, Fig.~\ref{minkowski_ours} for South-Russian sandstone, respectively, and are in good agreement. We also provide permeability comparison for all samples under consideration (Fig.~\ref{permeability_berea} --- Berea,
Fig.~\ref{permeability_ketton} --- Ketton, Fig.~\ref{permeability_ours} --- in-house sample).


%% file: sections_phys_rev/5_conclusion.tex
\section{Conclusion}
\label{conclusion}

In this paper, we proposed a new model for the generation of three-dimensional porous media. The novelty of our work is in the ability of generation 3D volumes from a 2D central slice. We showed that our model could be efficiently used in the reconstruction. For this purpose, we compared permeability, porosity, and other important functionals of real and synthetic data. Potentially, such a kind of model can be applied to seismic and geological data. Another possible application is texture generation.

In future work, it would be interesting to study other topological properties of generated samples, and also include such features as porosity into the loss directly. Another possible direction of research is the application of sparse 3D convolutions\cite{notchenko2017large} or generative models with a particular convolutional structure\cite{LCMICLR} to generate rock samples with high resolution.